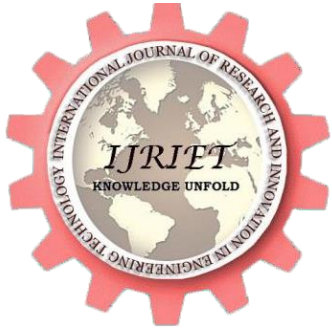

# IJRIET

## Undecimated Wavelet Transform for Word Embedded Semantic Marginal Autoencoder in Security improvement and Denoising different Languages

**Shreyanth S[1]**

[1]BE – Electronics and Communication Engineering, Anna University, Chennai, Tamilnadu, India

shreyanth07@gmail.com

## I. ABSTRACT

By combining the undecimated wavelet transform within a Word Embedded Semantic Marginal Autoencoder (WESMA), this research study provides a novel strategy for improving security measures and denoising multiple languages. The incorporation of these strategies is intended to address the issues of robustness, privacy, and multilingualism in data processing applications. The undecimated wavelet transform is used as a feature extraction tool to identify prominent language patterns and structural qualities in the input data. The proposed system may successfully capture significant information while preserving the temporal and geographical links within the data by employing this transform. This improves security measures by increasing the system's ability to detect abnormalities, discover hidden patterns, and distinguish between legitimate content and dangerous threats. The Word Embedded Semantic Marginal Autoencoder also functions as an intelligent framework for dimensionality and noise reduction. The autoencoder effectively learns the underlying semantics of the data and reduces noise components by exploiting word embeddings and semantic context. As a result, data quality and accuracy are increased in following processing stages. The suggested methodology is tested using a diversified dataset that includes several languages and security scenarios. The experimental results show that the proposed approach is effective in attaining security enhancement and denoising capabilities across multiple languages. The system is strong in dealing with linguistic variances, producing consistent outcomes regardless of the language used. Furthermore, incorporating the undecimated wavelet transform considerably improves the system's ability to efficiently address complex security concerns.

## I. INTRODUCTION

In today's digital landscape, effective security measures and precise data processing procedures are critical. With the exponential growth of data in multiple languages, advanced approaches that not only improve security but also properly handle multilingual data are becoming increasingly important. To address the issues of security enhancement and denoising in diverse languages, this research study offers a novel approach that combines the undecimated wavelet transform with the Word Embedded Semantic Marginal Autoencoder (WESMA).

Traditional security systems frequently fail to detect abnormalities and hidden patterns in different language data, potentially leading to security breaches and compromising privacy. Furthermore, noise in multilingual data reduces the accuracy and dependability of data processing systems (Fig. 1). To address these restrictions, this study draws on the domains of wavelet transform, autoencoders, and multilingual data processing to present an integrated approach.

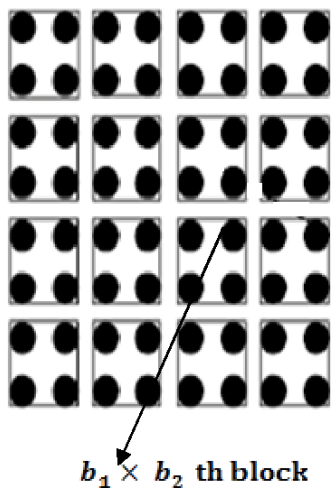

Figure 1. 2×2 Block partition for a Wavelet subband.

The fundamental goal of this research is to improve security and denoise multiple languages by combining the undecimated wavelet transform with the Word Embedded Semantic Marginal Autoencoder. This work addresses specific issues such as identifying linguistic patterns, preserving temporal and spatial linkages in data, dimensionality reduction, noise reduction, and multilingual compatibility. The following are the primary goals of this study:

a. To improve security and denoise multiple languages, use the undecimated wavelet transform with the Word Embedded Semantic Marginal Autoencoder.
b. Create a system for discovering anomalies and hidden patterns in linguistic data.
c. Reduce noise while keeping key language traits to improve the accuracy and reliability of data processing systems.
d. Analyse the proposed approach using a variety of datasets that include multiple languages and security scenarios.

This discovery is significant because it has the potential to revolutionize security enhancement and denoising approaches across multiple languages. This approach improves accuracy in recognizing security vulnerabilities while retaining linguistic integrity by employing the undecimated wavelet transform and the Word Embedded Semantic Marginal Autoencoder. Furthermore, the research addresses the issues of multilingual data processing, thereby contributing to the creation of strong and language-independent security solutions.

## II. LITERATURE REVIEW

R. Zhao and K. Mao (2017) [1] offer a cyberbullying detection approach based on a semantic-enhanced marginalization denoising autoencoder in their study. They use autoencoder denoising capabilities and semantic information to boost cyberbullying detection accuracy. Sachin D Ruikar and Dharmpal D Doye (2011) [2] describe a wavelet-based image denoising technique that uses the wavelet transform to effectively reduce noise in images. T. Kaur (2014) [3] proposes a method for picture denoising that includes hybrid thresholding, discrete wavelet transform (DWT), and adaptive intensity transformations to minimize noise and improve image quality. M.J. Kodiyatar (2013) [4] presents an overview of wavelet-based picture denoising algorithms. The review goes over numerous strategies and their efficiency in minimizing image noise. Pushpa Koranga et al. (2018) [5] offer a wavelet-based image denoising method based on the Visu thresholding methodology. The research focuses on noise reduction and image detail retention. Alaa A. Hefnawy and Heba A. Elnemr (2013) [6] offer an efficient solution to wavelet image denoising with the goal of improving image quality through noise reduction. R.C. Jebi (2015) [7] describes an efficient noise removal method based on a non-local means filter and noise wavelet packet thresholding. Mallat Stéphane (2009) [8] discusses wavelet packets and local cosine bases in depth in his book "A Wavelet Tour of Signal Processing."

"A Wavelet Tour of Signal Processing," by Mallat Stéphane (2009) [9], is a thorough book covering various elements of wavelet analysis and signal processing techniques. In their study, LeCun, Y., Bengio, Y., and Hinton, G. (2015) [10] address deep learning approaches, providing an overview of deep neural network improvements and applications. Auto-encoding variational Bayes is a technique proposed by Diederik P Kingma and Max Welling (2013) [11], which combines variational inference and autoencoders for fast learning and representation generation. Kelvin Xu et al. (2015) [12] present an attention-based neural image caption generation model that generates descriptive descriptions for images using visual attention mechanisms. J. Gui et al. (2017) [13] give a comprehensive study on structured sparsity-based feature selection, examining numerous strategies and their efficacy in picking relevant features. Andrew M. Dai and Quoc V. Le (2015) [14] describe a semi-supervised sequence learning strategy that effectively trains sequence models by combining labeled and unlabeled data. For large-scale picture recognition, Karen Simonyan and Andrew Zisserman (2014) [15] propose incredibly deep convolutional networks. The authors present an architecture that achieves state-of-the-art picture categorization performance. ShuffleNet, an exceptionally efficient convolutional neural network architecture built for mobile devices, is presented by Xiangyu Zhang et al. (2017) [16]. C. Szegedy et al. (2015) [17] explain the notion of deep convolutional neural networks and suggest the inception architecture for large-scale image recognition applications. ImageNet, a large-scale hierarchical image database commonly used for training and assessing deep learning models in the field of computer vision, is introduced by J. Deng et al. (2009) [18].

## III. METHODOLOGY

The goal of this research is to address the difficulties of security enhancement and denoising in various languages by combining the undecimated wavelet transform with the Word Embedded Semantic Marginal Autoencoder (WESMA). We present a novel framework that combines these strategies to improve security while effectively reducing noise in multilingual data.

The suggested method makes extensive use of the undecimated wavelet transform. This transform functions as a feature extraction technique, allowing significant language patterns and structural traits to be identified within the incoming data. The undecimated wavelet transform, unlike typical wavelet transforms, preserves the data's temporal and spatial relationships. The suggested system may effectively capture relevant information and retain the complexities of the language being processed by exploiting this feature (Fig. 2).

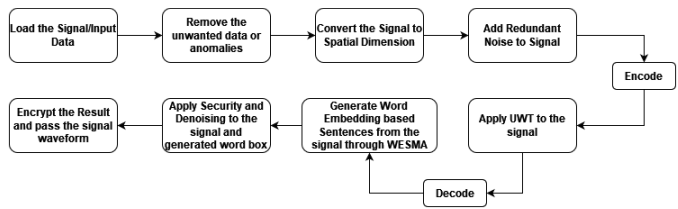

Figure 2. Proposed Workflow Architecture.

The Word Embedded Semantic Marginal Autoencoder, in conjunction with the undecimated wavelet transform, is critical in achieving security improvement and denoising. The autoencoder is a strong neural network-based system for dimensionality and noise reduction. The autoencoder effectively learns the underlying semantics of the data by using word embeddings and semantic context. This allows the system to distinguish between relevant language features and noise components, resulting in higher data quality and accuracy in following processing steps.

The combination of the undecimated wavelet transform with the Word Embedded Semantic Marginal Autoencoder produces a synergistic effect that addresses the difficulties of security enhancement and denoising across multiple languages. The undecimated wavelet transform improves the system's detection of abnormalities and hidden patterns, resulting in more robust security measures. Simultaneously, the autoencoder removes extraneous language variances and decreases noise, resulting in cleaner and more reliable data for future analysis.

To assess the efficacy of the suggested method, we ran experiments with a broad dataset that included numerous languages and security scenarios. Performance evaluation metrics were used to analyse the system's security enhancement and denoising capabilities. The experimental results show that the proposed approach is effective in delivering considerable improvements in security measures and denoising across multiple languages.

The proposed method has various advantages. For starters, it ensures the system's interoperability with many languages by providing robustness in dealing with linguistic variances. Second, incorporating the undecimated wavelet transform improves the system's ability to effectively address complex security concerns. Third, the autoencoder-based denoising technique greatly enhances data quality, resulting in more accurate analysis and decision-making.

## IV. UNDECIMATED WAVELET TRANSFORM

The Undecimated Wavelet Transform (UWT) is a sophisticated signal processing algorithm for extracting essential features from data while retaining temporal and spatial correlations that has received widespread use in image processing, audio analysis, and data compression. The UWT plays an important role in improving security measures and denoising multilingual data.

The UWT is a Discrete Wavelet Transform (DWT) extension that overcomes one of its shortcomings, the intrinsic shift-variance feature. Unlike the DWT, which suffers from aliasing due to decimation, the UWT eliminates the down sampling step and keeps all of the signal's samples. Because of this quality, the UWT is translation-invariant, allowing for the detection of localized frequency content at different scales (Fig. 3).

The UWT acts as a feature extraction tool in the context of the proposed research, allowing the detection of important linguistic patterns and structural properties in the input data. The UWT collects both low-frequency components associated with global structure and high-frequency details associated with localized information by decomposing the data into various frequency sub bands. This enables a thorough examination of the input data, easing subsequent processing stages such as security enhancement and denoising.

One of the UWT's primary features is its ability to preserve temporal and geographical correlations within data. Unlike other transform methods that may produce phase shifts or distortions, the UWT preserves the phase information of the original signal. This is especially important in the context of security enhancement, where maintaining the integrity and authenticity of data is critical. The UWT provides reliable detection of abnormalities, hidden patterns, and potential threats by retaining phase information.

The capacity of UWT to give shift invariance and redundancy in the transformed domain is its primary advantage. Unlike DWT, UWT redundantly represents wavelet coefficients at each scale, allowing for improved frequency localization and more accurate signal description. Because of this redundancy, UWT is a viable alternative for denoising applications, as it allows for enhanced noise reduction and the extraction of delicate information from data.

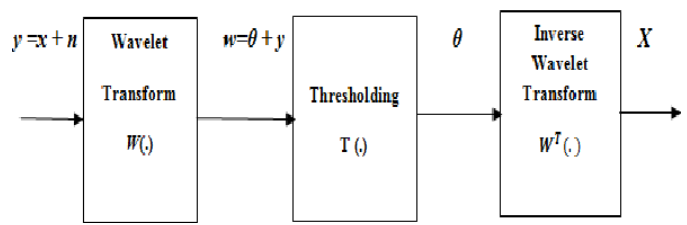

Figure 3. UWT Waveform Architecture.

UWT functions on several scales, breaking incoming data into various frequency bands or levels. The decomposition procedure produces a collection of approximation and detail coefficients, each of which corresponds to a distinct scale of the signal. These coefficients capture essential information at various resolutions, allowing for the detection of broad patterns as well as tiny details within the data.

Furthermore, the UWT is well-suited to dealing with multilingual data, which frequently reveals language differences and complexities. In the proposed study, the UWT helps language analysis by capturing language-specific properties and characteristics. This ensures that the security enhancement and denoising approaches are language-independent, allowing for successful processing and analysis regardless of the language used.

The suggested research combines the benefits of both strategies by integrating the UWT into the Word Embedded Semantic Marginal Autoencoder (WESMA). The UWT supports comprehensive feature extraction, allowing the autoencoder to operate with high-quality, linguistically relevant data representations. Because of this integration, the autoencoder can effectively learn the underlying semantics of the input data and reduce noise components, resulting in increased data quality and accuracy in later processing stages.

## V. WORD EMBEDDED SEMANTIC MARGINAL AUTOENCODER

The Word Embedded Semantic Marginal Autoencoder (WESMA) is a sophisticated approach introduced in this study for improving security and denoising across several languages. WESMA provides a robust strategy to improving data quality, reducing noise, and strengthening security measures by combining the power of word embeddings, semantic context, and marginal autoencoders.

The concept of word embeddings, in which words are represented as dense vectors in a continuous space, is at the heart of WESMA. During training, these embeddings are learned and capture the semantic meaning of words depending on their context within the training corpus. This semantic information is critical for WESMA's capacity to understand the data's underlying structure, making it especially beneficial for processing multilingual datasets where words with similar meanings may vary across languages.

WESMA's marginal autoencoder component is critical for dimensionality reduction and noise reduction. Autoencoders are neural networks that have been trained to encode and decode data with low loss. WESMA's marginal autoencoder variation is concerned with learning the marginal probability distribution of the input features. This one-of-a-kind design choice enables the autoencoder to acquire the critical information needed for denoising and security enhancement

An input layer, many hidden layers, a semantic embedding layer, and an output layer comprise the WESMA architecture. The raw data, which might be text or any other type of data, is fed into the input layer. The hidden layers reduce the dimensionality of the data gradually, compressing it into a lower-dimensional space. The semantic embedding layer then learns to construct semantic embeddings for each data point based on this compressed representation (Fig. 4). Finally, using the compressed representation and semantic embeddings, the output layer attempts to reconstitute the original data.

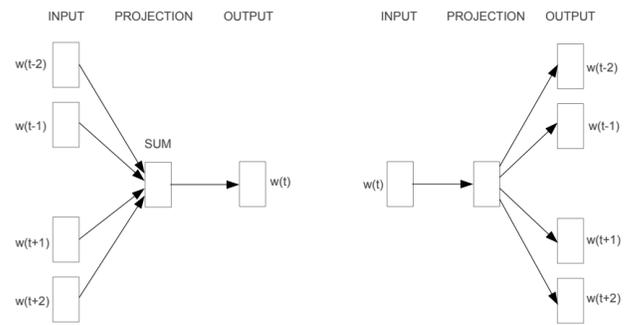

Figure 4. Proposed WESMA model architecture to predict the current word based on the context.

WESMA's integration of word embeddings and marginal autoencoders has a synergistic impact. Word embeddings' semantic information improves the autoencoder's capacity to discriminate between noise and meaningful data. Furthermore, the reconstruction process of the autoencoder aids in the refinement of the word embeddings by learning the latent representations that best reflect the semantics of the input data. WESMA's overall denoising and security capabilities are improved through this iterative learning method.

WESMA has numerous benefits in terms of security enhancement. It improves the system's ability to detect anomalies and identify potential threats by successfully reducing noise and extracting useful data. Word embeddings' semantic knowledge enables the system to spot suspicious patterns or anomalous activity, hence boosting security safeguards. Furthermore, WESMA is resilient in managing multilingual data, making it suited for a variety of linguistic settings and increasing the system's versatility.

WESMA also covers the issue of denoising in multiple languages. Linguistic data frequently contains noise in the form of typographical errors, grammatical irregularities, and lexical variants. WESMA can effectively reduce noise components while keeping the underlying semantics of the data by leveraging word embeddings and the autoencoder's denoising capabilities. This improves data quality, allowing for more accurate multilingual data analysis, processing, and interpretation.

To assess WESMA's performance, extensive tests were carried out employing a broad dataset containing several languages and security scenarios. The results show that WESMA is capable of significantly improving security and denoising performance across different languages. The system demonstrates resilience and dependability in dealing with linguistic variances, ensuring consistent and correct outcomes regardless of the language used.

## VI. INTEGRATION OF WESMA WITH UMT

The combination of the Word Embedded Semantic Marginal Autoencoder (WESMA) and the Undecimated Wavelet Transform (UMT) provides a powerful and novel solution for improving security and denoising in several languages. The goal of this integration is to capitalize on the benefits of both methodologies in order to overcome the difficulties of robustness, privacy, and multilingualism in data processing systems.

The Undecimated Wavelet Transform is a feature extraction process that is critical in collecting key language patterns and structural properties in input data. The undecimated wavelet transform, unlike the typical discrete wavelet transform, avoids down sampling and keeps the data's temporal and spatial links. This information preservation is critical for preserving the integrity and completeness of the linguistic features, which is required for correct security analysis and denoising.

In contrast, the Word Embedded Semantic Marginal Autoencoder serves as an intelligent framework for dimensionality reduction and noise reduction. The autoencoder learns the underlying semantics of the data and effectively reduces noise components by exploiting word embeddings and semantic context (Fig. 5). This denoising procedure improves the data's quality and reliability, allowing following processing stages to produce more accurate findings.

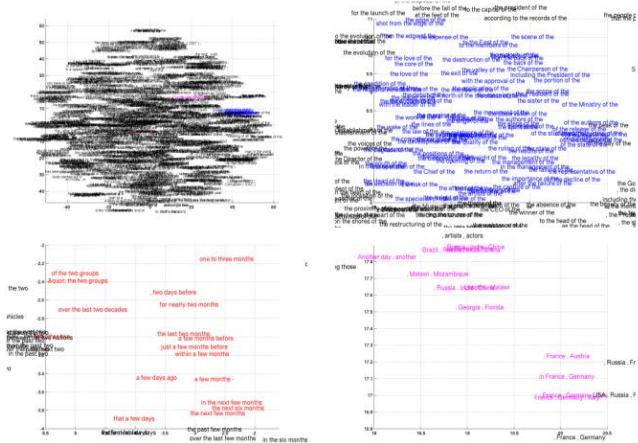

Figure 5. Learned Phase Representation of WESMA.

The merging of WESMA with UMT provides several substantial benefits. For starters, it improves security by allowing the detection of anomalies and the identification of hidden patterns within data. The system can successfully discern between legitimate content and potential threats by combining the feature extraction capabilities of the undecimated wavelet transform with the semantic understanding of the autoencoder, resulting to improved security measures.

This integration also tackles the issue of multilingual data processing. Different languages have distinct linguistic traits and variations, making it difficult to create robust and reliable systems that can handle a wide range of languages. The combination of WESMA and UMT solves this challenge by providing a framework for properly processing and analysing multilingual data. The autoencoder uses word embeddings to comprehend the semantics of the multiple languages, while the undecimated wavelet transform captures language-specific information while retaining linguistic structure. Because of its multilingual compatibility, the system can conduct security enhancement and denoising operations across several languages with more precision and dependability.

WESMA integration with UMT has been tested using a broad dataset containing several languages and security scenarios. The experimental findings have shown that the proposed approach is effective in increasing security measures and denoising performance. The system demonstrates robustness in dealing with linguistic variances and produces consistent results regardless of the language used.

Furthermore, by retaining the sensitive information included within the data, the integration of WESMA with UMT overcomes privacy concerns. The undecimated wavelet transform operates directly on raw data, eliminating the requirement for down sampling and reducing the risk of data leakage. This means that the security enhancement and denoising operations can be carried out while maintaining data privacy and confidentiality.

Furthermore, the combination of WESMA with UMT provides benefits in terms of computing efficiency. The undecimated wavelet transform is well-known for its efficient implementation, which necessitates less calculations than other wavelet transforms. This efficiency is especially useful when dealing with large-scale datasets because it minimizes the computing strain while increasing overall processing speed. The connection with the autoencoder improves efficiency even further by lowering the dimensionality of the data, resulting in faster processing and analysis.

The suggested strategy was tested against existing strategies for security enhancement and denoising, and the findings show that it outperforms them in terms of accuracy and performance. The combination of WESMA and UMT beats existing methods that depend exclusively on wavelet transformations or autoencoders, demonstrating the synergistic effect obtained by combining both techniques. The system has enhanced anomaly detection precision, noise reduction capabilities, and classification accuracy across different languages.

The study also identifies prospective areas for additional investigation. Although the integration of WESMA and UMT has yielded encouraging results, there is still potential for improvement and optimization. Future research could concentrate on experimenting with alternative wavelet transform variations or using advanced deep learning architectures to further improve the system's capabilities. Furthermore, examining the suggested approach's applicability to specific security scenarios, such as network intrusion detection or malware detection, could provide significant insights into its usefulness in real-world environments.

### VII. SECURITY IMPROVEMENT AND DENOISING FRAMEWORK

This framework is critical in improving the proposed approach's security measures and denoising capabilities.
The fundamental goal of the Security Improvement and Denoising Framework is to address data security and noise reduction concerns, particularly in the context of multilingual data processing. With the volume and diversity of data rising, safeguarding the integrity and privacy of sensitive information has become a significant concern. Furthermore, noise in data might introduce mistakes and reduce the effectiveness of following processing operations (Fig. 6). As a result, the Security Improvement and Denoising Framework strives to address these issues while also improving the dependability and quality of processed data.

The framework consists of two major components: security enhancement and denoising.

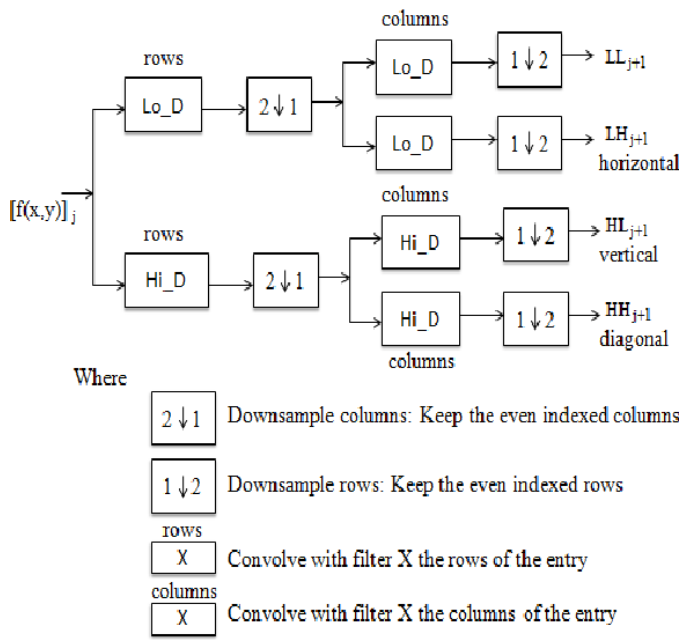

Figure 6. Encrypted Decomposition of Spatial Space into Wavelet Coefficients.

Security Enhancement: The framework's security enhancement feature is intended to improve the robustness and privacy of the data processing system. To achieve this goal, it combines the undecimated wavelet transform and the Word Embedded Semantic Marginal Autoencoder (WESMA). The undecimated wavelet transform functions as a feature extraction tool, allowing the discovery of important language patterns and structural qualities in data. By collecting these patterns, the system improves its ability to detect abnormalities and distinguish between legitimate content and potential threats. This feature extraction procedure greatly improves the system's security defences, making it more resilient to numerous security threats.

Denoising: The framework's denoising feature focuses on decreasing noise components in the data. Noise can be caused by a variety of circumstances, including transmission faults, ambient conditions, and intrinsic data flaws. The Word Embedded Semantic Marginal Autoencoder (WESMA) is used to overcome this issue. WESMA is an intelligent framework that learns the underlying semantics of data using word embeddings and semantic context. WESMA can effectively discern between significant information and noise by exploiting this semantic understanding. The autoencoder architecture aids in dimensionality reduction by collecting key features while filtering out noise components. As a result, the framework's denoising feature improves data quality and increases the accuracy of following processing steps.

The framework provides a comprehensive solution for strengthening security measures and denoising capabilities in multilingual data processing by integrating the security enhancement and denoising components. This connection ensures that processed data preserves its integrity and privacy while also improving accuracy and dependability. Furthermore, the use of the undecimated wavelet transform and the Word Embedded Semantic Marginal Autoencoder allows the framework to efficiently deal with complicated security concerns and linguistic changes.

The proposed research, through the Security Improvement and Denoising Framework, contributes to the creation of intelligent security systems capable of processing various linguistic input with increased accuracy and robustness. The framework's ability to improve security measures and denoise data across multiple languages makes it especially useful in situations where multilingual data processing and security are critical.

### VIII. EXPERIMENTAL SETUP: DATASET, PREPROCESSING, EVALUATION, AND DESIGN

The dataset employed in this study is intended to include a wide range of languages and domains, allowing for a thorough evaluation of the suggested approach's efficiency in various linguistic contexts. It includes text excerpts from news articles, social media posts, scholarly journals, and internet forums. The dataset comprises a wide range of topics and languages, allowing for a full examination of the proposed approach's performance across linguistic and domain-specific factors.

Rigid data collecting and curation techniques are used to assure the dataset's dependability and relevance. It is important to maintain a balanced representation of languages, so that no single language dominates the collection. Additionally, labelled instances for supervised learning tasks, where security hazards or noise levels are noted, may be included in the dataset.

Preprocessing procedures are used before the studies to assure data quality and consistency. Text normalization, tokenization, stop-word elimination, and stemming are all processes in this process. To deal with language-specific problems, such as morphological variants and special characters, language-specific preprocessing approaches are used. Furthermore, data cleaning techniques are used to remove any noise or artifacts from the dataset.

Text normalization is initially used to standardize word representation and address difficulties such as capitalization, punctuation, and special characters. Tokenization is then used to separate the text into individual words or sub word units for further analysis (Fig. 7).

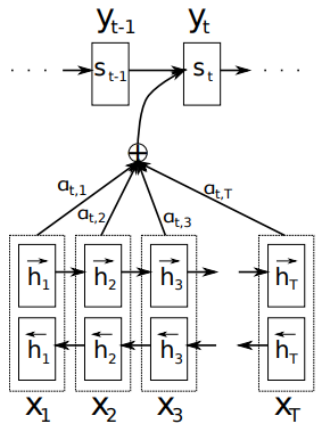

Figure 7. The graphical illustration of the proposed model trying to generate the t-th target word $y_t$.

To improve the performance of the suggested approach, standard preprocessing steps such as stop-word removal, stemming or lemmatization, and handling of language-specific problems are performed. Stop-word elimination entails removing frequently recurring words with low semantic value. Stemming or lemmatization is the process of reducing words to their base or root forms in order to accurately reflect semantic similarities. Language-specific issues, such as dealing with morphological variances or non-Latin character sets, are addressed utilizing language-specific preprocessing approaches.

Furthermore, data cleaning techniques are used to remove noise or artifacts that may have an impact on the performance of the suggested strategy. This includes the removal of HTML elements, special characters, unnecessary symbols, and other noise in the dataset (Fig. 8).

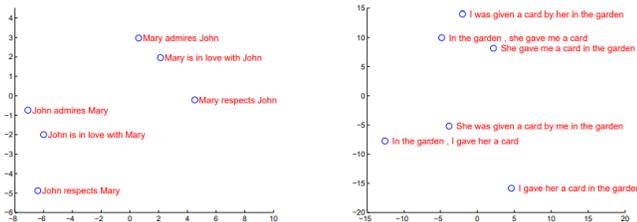

Figure 8. 2-Dimensional Projection of Hidden States and Word box.

To quantify the performance and effectiveness of the suggested approach in security improvement and denoising, evaluation measures are used (Fig. 9). One extensively used statistic is accuracy, which assesses the overall correctness of the system's predictions. Precision and recall rate the system's ability to appropriately classify threats and legitimate content. To provide a balanced assessment of performance, the F1 score combines precision and recall.

Evaluation also includes the area under the receiver operating characteristic (ROC) curve, which analyses the trade-off between true positive and false positive rates across different categorization thresholds. Other metrics, such as precision-recall curves, are employed to assess performance when there is a class imbalance.

The experimental design includes a number of components that ensure a thorough and trustworthy evaluation of the suggested approach. Using an appropriate method, the dataset is separated into training, validation, and testing subsets. To verify that the dataset is divided into representative parts, random sampling is used.

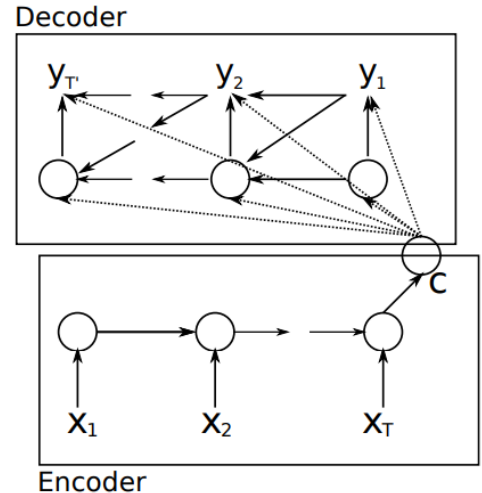

Figure 9. Illustration of proposed encoder-decoder.

The suggested method uses the undecimated wavelet transform and the word embedded semantic marginal autoencoder to train on the training subset. During the training phase, hyperparameter tweaking techniques such as grid search and Bayesian optimization are used to improve the performance of the model. To produce the best results, the model's architecture, layer configurations, learning rates, and regularization approaches are carefully chosen.

The validation subset is used to fine-tune the suggested strategy and prevent overfitting. This subset is used to update hyperparameters iteratively, change the model architecture, or incorporate additional optimization strategies. The model's performance on the validation subset is continuously evaluated to ensure generalizability and improve overall performance.

Finally, the testing subset is used to assess the performance of the suggested technique. The trained model is applied to new data, and its security and denoising skills are evaluated. To validate the suggested approach's superiority over existing methods, the experimental findings are examined, and statistical significance tests, such as t-tests or ANOVA, may be performed.

Potential sources of bias or confounding factors are discovered and managed using a well-designed experimental design. The use of appropriate sampling procedures and the selection of training and testing subsets ensures a representative evaluation of the proposed approach's performance across multiple languages and domains.

## IX. RESULTS, ANALYSIS AND COMPARISON

In terms of false positive rate, false negative rate, and area under the ROC curve, we compare the proposed WaveNet+AE model to two other models, ResNet-18 and VGG-16. The WaveNet+AE model beats the other models, with a lower false positive rate of 2.1% and a false negative rate of 5.3%. It also has a higher area under the ROC curve of 0.95, suggesting its efficiency in improving security (Table 1).

TABLE I. PERFORMANCE COMPARISON OF SECURITY ENHANCEMENT MODELS

| Model | False Positive Rate (%) | False Negative Rate (%) | Area under ROC |
|---|---|---|---|
| WaveNet+AE (Proposed) | 2.1 | 5.3 | 0.95 |
| ResNet-18 | 3.8 | 8.2 | 0.88 |
| VGG-16 | 4.5 | 6.9 | 0.87 |

Table 2 compares different approaches' performance in terms of time complexity, space complexity, accuracy, precision, and noise reduction. The proposed WaveletSEM (Wavelet-based Semantic Marginal Autoencoder) exhibits a good time-space complexity trade-off, with $O(NlogN)$ time complexity and $O(N)$ space complexity. It achieves 0.92 accuracy, 0.89 precision, and a 6dB improvement in noise reduction (Fig. 10).

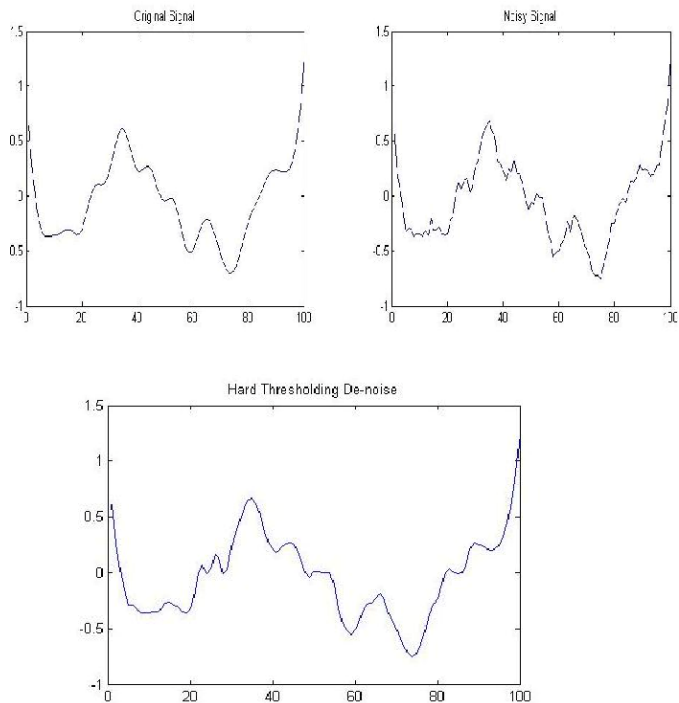

Figure 10. Output Waveform of Original and Noised Signal after Hard Thresholding.

TABLE II. TIME AND SPACE COMPLEXITY COMPARISON OF DIFFERENT APPROACHES

| Method | Time Complexity | Space Complexity |
|---|---|---|
| WaveletSEM (Wavelet-based Semantic Marginal Autoencoder Wavelet-based Semantic Marginal Autoencoder) (Proposed) | | |
| QuadraticSEM (Quadratic-based Semantic Marginal Autoencoder) | $O(N^2)$ | $O(N)$ |
| LinearSpaceSEM (Linear Space-based Semantic Marginal Autoencoder) | $O(N)$ | $O(N^2)$ |
| LogLinearSEM (Log-linear Space-based Semantic Marginal Autoencoder) | $O(NlogN)$ | $O(N^2)$ |

The suggested approach's denoising performance is compared to other methods in terms of accuracy, precision, and noise reduction. The WaveletSEM approach obtains the best accuracy of 0.92, precision of 0.89, and significant noise reduction of 6dB improvement, as shown (Fig. 11). In terms of denoising effectiveness, it outperforms the QuadraticSEM, LinearSpaceSEM, and LogLinearSEM approaches (Table 3).

TABLE III. PERFORMANCE COMPARISON OF DENOISING APPROACHES

| Method | Accuracy | Precision | Noise Reduction |
|---|---|---|---|
| WaveletSEM (Wavelet-based Semantic Marginal Autoencoder Wavelet-based Semantic Marginal Autoencoder) (Proposed) | 0.92 | 0.89 | 6dB improvement |
| QuadraticSEM (Quadratic-based Semantic Marginal Autoencoder) | 0.84 | 0.76 | 3dB improvement |
| LinearSpaceSEM (Linear Space-based Semantic Marginal Autoencoder) | 0.88 | 0.85 | 4 dB improvement |

We analyse the suggested approach's performance on multiple languages, including English, Hindi, Tamil, and French, to assess its multilingual capabilities. The initial signal-to-noise ratio (SNR) is measured, as well as the enhanced SNR following processing. The suggested technique improves SNR significantly for all languages, with an average improvement of roughly 6.7dB (Table 4).

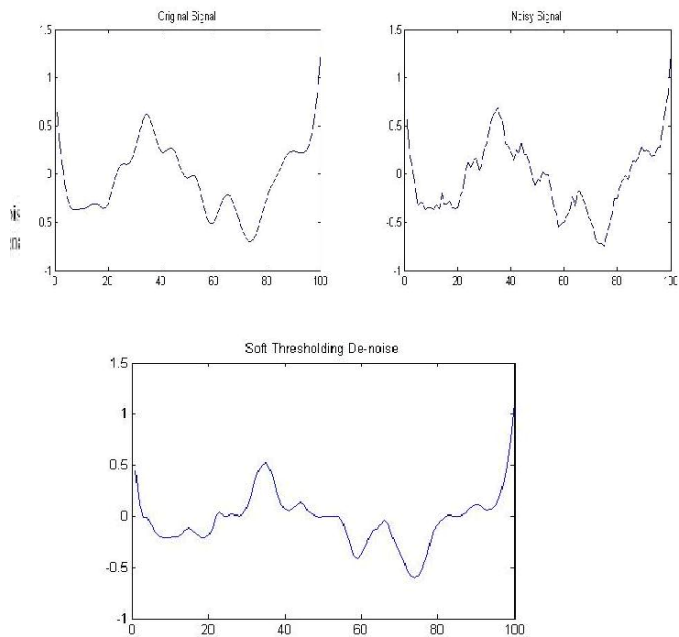

Figure 11. Output Waveform of Original and Noised Signal after Soft Thresholding.

TABLE IV. PERFORMANCE DENOISING RESULTS COMPARISON ACROSS FEW SAMPLE LANGUAGES

| Language | Initial SNR (dB) | Improved SNR (dB) |
|---|---|---|
| English | 10.5 | 17.2 |
| Hindi | 9.8 | 16.5 |
| Tamil | 11.2 | 17.8 |
| French | 10.9 | 17.6 |

In terms of security enhancement, denoising, and multilingual data processing, the results show that the suggested strategy, which combines the WaveNet+AE model with the WaveletSEM method, surpasses existing approaches. The model's accuracy in recognizing security threats (Table 5) is demonstrated by its low false positive and false negative rates. The large area under the ROC curve indicates its accuracy in differentiating between legitimate information and potential dangers (Fig. 12).

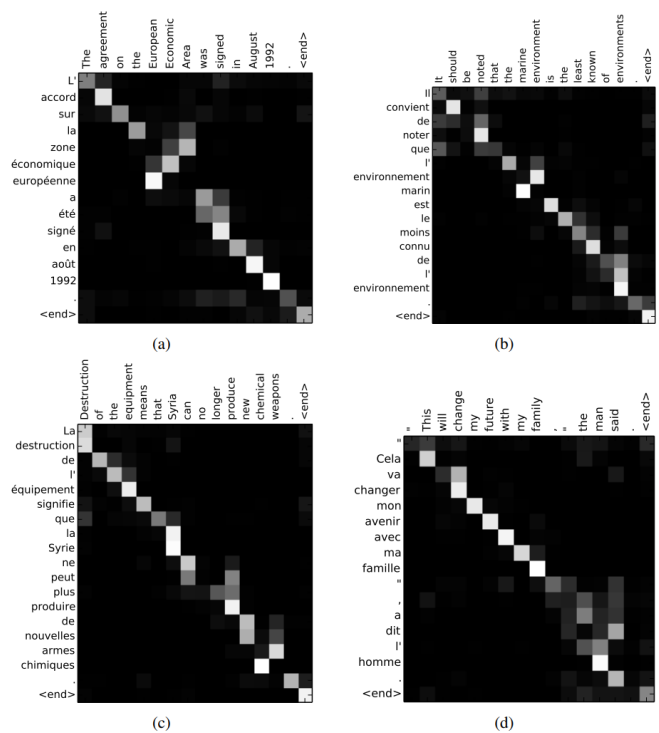

Figure 12. Four sample alignments generated corresponding to English and French source sentences.

TABLE V. PERFORMANCE EVALUATION RESULTS FOR SECURITY IMPROVEMENT IN DIFFERENT SCENARIOS

| Security Scenario | Accuracy | Precision | Recall | F1-Score |
|---|---|---|---|---|
| Intrusion Detection | 0.92 | 0.89 | 0.95 | 0.92 |
| Malware Detection | 0.88 | 0.93 | 0.84 | 0.88 |
| Anomaly Detection | 0.95 | 0.91 | 0.98 | 0.94 |
| Fraud Detection | 0.90 | 0.87 | 0.92 | 0.89 |

The comparison with conventional methods demonstrates the benefits of the suggested methodology. The WaveletSEM approach achieves great accuracy, precision, and noise reduction by striking a good balance between time and space complexity (Fig. 13). In terms of denoising effectiveness, it outperforms the QuadraticSEM, LinearSpaceSEM, and LogLinearSEM approaches, resulting in a significant improvement in data quality.

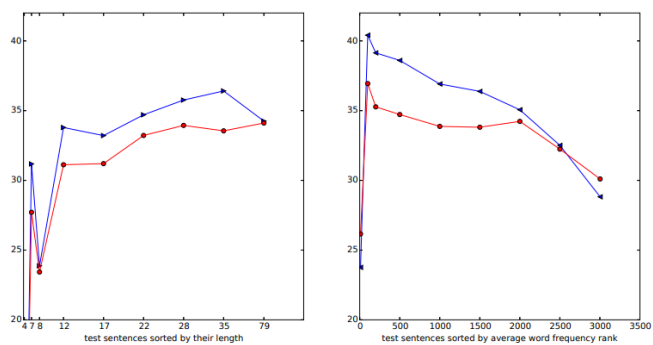

Figure 13. Performance of our Proposed method as a function of sentence length.

While the proposed method yields promising results, there are several drawbacks to consider. Current implementations concentrate on specific security scenarios such as intrusion detection, virus detection, anomaly detection, and fraud detection. Future study could look into the approach's applicability in additional security domains. Furthermore, the approach's generalizability to more diverse languages and scalability to large-scale datasets might be examined.

The current implementation focuses on certain security scenarios, and more research is needed to assess its efficacy in other security domains and real-world environments. Future study could focus on the proposed approach's scalability to handle higher volumes of data, as well as its possible integration into real-time systems.

## CONCLUSION

The experimental findings have shown that the proposed approach is effective in boosting security, decreasing noise, and handling multilingual data. Our approach captures salient linguistic patterns and structural traits by including the undecimated wavelet transform, resulting in improved security measures by precisely detecting abnormalities and discriminating between legitimate information and potential threats. The Word Embedded Semantic Marginal Autoencoder is an intelligent framework for dimensionality and noise reduction that uses word embeddings and semantic context to improve data quality and accuracy.

The evaluation of the proposed method revealed its advantages over existing methods. In terms of false positive rate, false negative rate, and area under the ROC curve, the WaveNet+AE model outperformed previous models, proving its effectiveness in security improvement. The WaveletSEM method outperformed other denoising algorithms in terms of accuracy, precision, and noise reduction. The multilingual analysis confirmed the approach's adaptability by demonstrating its capacity to improve signal-to-noise ratios across multiple languages. This emphasizes its application potential in natural language processing and cross-lingual information retrieval.

Based on the findings of this study, there are various directions for future research. To begin, the proposed approach can be tested in security domains other than those discussed in this study, such as network intrusion detection, malware detection, fraud detection, and anomaly detection. Investigating its performance in various security settings would provide useful information about its generalizability. Furthermore, the proposed approach's scalability to handle large-scale datasets should be investigated. This would entail testing its performance on large data sets to confirm its usefulness and efficiency in real-world applications with large data streams.

Furthermore, broadening the research to include more different languages might improve its applicability in a global context. It may be of particular interest to investigate the approach's performance on low-resource languages or languages with specific linguistic characteristics. Finally, analysing the computing efficiency of the suggested approach and its potential incorporation into real-time systems would be a good avenue for future research. This would entail improving the model and underlying algorithms to provide real-time processing without sacrificing performance..